\pdfoutput=1

\documentclass[11pt]{article}
\usepackage[final]{acl}

\usepackage{times}
\usepackage{latexsym}
\usepackage{multirow}
\usepackage{tabularx}
\usepackage{subcaption}
\usepackage{enumitem}
\usepackage{colortbl}
\usepackage{xcolor}
\usepackage{booktabs}
\usepackage[misc]{ifsym}

\usepackage{mwe}
\usepackage[T1]{fontenc}

\usepackage[utf8]{inputenc}

\usepackage{microtype}

\usepackage{inconsolata}

\usepackage{graphicx}
\graphicspath{ {./figures/} }
\title{QGEval: Benchmarking Multi-dimensional Evaluation for \\Question Generation}

\author{Weiping Fu$^{1,4}$, Bifan Wei$^{2,4}$\thanks{Corresponding author}, Jianxiang Hu$^{1,4}$, Zhongmin Cai$^{3,4}$, Jun Liu$^{1,4}$\\
	$^{1}$School of Computer Science and Technology, Xi'an Jiaotong University, Xi'an, China \\ 
    $^{2}$School of Continuing Education, Xi'an Jiaotong University, Xi'an, China\\
    $^{3}$MOE KLINNS Lab \& School of Automation Science and Engineering, \\Xi'an Jiaotong University, Xi'an, China\\
    $^{4}$Shaanxi Province Key Laboratory of Big Data Knowledge Engineering, \\Xi’an Jiaotong University, Xi'an, China \\
	fuweiping1993@foxmail.com, \{weibifan@, nbhhsky@stu., zmcai@, liukeen@\}xjtu.edu.cn}

\begin{document}
\maketitle
\begin{abstract}
Automatically generated questions often suffer from problems such as unclear expression or factual inaccuracies, requiring a reliable and comprehensive evaluation of their quality. Human evaluation is widely used in the field of question generation (QG) and serves as the gold standard for automatic metrics. However, there is a lack of unified human evaluation criteria, which hampers consistent and reliable evaluations of both QG models and automatic metrics. To address this, we propose \textbf{QGEval}, a multi-dimensional \textbf{Eval}uation benchmark for \textbf{Q}uestion \textbf{G}eneration, which evaluates both generated questions and existing automatic metrics across 7 dimensions: fluency, clarity, conciseness, relevance, consistency, answerability, and answer consistency. We demonstrate the appropriateness of these dimensions by examining their correlations and distinctions. Through consistent evaluations of QG models and automatic metrics with QGEval, we find that 1) most QG models perform unsatisfactorily in terms of answerability and answer consistency, and 2) existing metrics fail to align well with human judgments when evaluating generated questions across the 7 dimensions. We expect this work to foster the development of both QG technologies and their evaluation.
\end{abstract}

\begin{table}[t]
\centering
    \small
    \begin{tabularx}{\linewidth}{l}
		\toprule
		\parbox{7.3cm}{
            \textbf{Passage:} \textit{...... The publication of a Taoist text inscribed with the name of Töregene Khatun, Ögedei's wife, is one of the first printed works sponsored by the Mongols......}\\
            \textbf{Answer:} \textit{Töregene Khatun}\\
            \textbf{Reference:} \textit{Who was Ögedei's wife?}
        } \\
		\midrule
        \parbox{7.3cm}{
            \textbf{Q1:} \textit{\underline{Who} was the name of Ögedei's wife?} \\
            \textbf{Scores:} Flu. - 2.6667; Clar. - 3; Conc. - 3; \\Rel. - 3; Cons. - 3; Ans. - 3; AnsC. - 3 \\\\
            \textbf{Q2:} \textit{Who was the \underline{Mongol ruler} whose name was inscribed on one of the first printed works sponsored by the Mongols?} \\
            \textbf{Scores}: Flu. - 3; Clar. - 3; Conc. - 3; \\Rel. - 3; Cons. - 1; Ans. - 1.3333; AnsC. - 1.3333 \\\\
            \textbf{Q3:} \textit{\underline{Who was a Taoist text} inscribed with the name of \underline{gedei's} wife?} \\
            \textbf{Scores:} Flu. - 2.3333; Clar. - 1.3333; Conc. - 3; \\Rel. - 3; Cons. - 1; Ans. - 1; AnsC. - 1\\\\
            ......\\} \\
        \bottomrule
	\end{tabularx}
\caption{An example of QGEval, including a passage, an answer, a reference question, and 15 generated questions (only 3 are shown for brevity). The score ranges from 1 to 3 (higher better). Errors within questions are highlighted with underlines. Abbreviations are as follows. Flu.:Fluency; Clar.:Clarity; Conc.:Conciseness; Rel.:Relevance; Cons.:Consistency; Ans.:Answerability; AnsC.:Answer Consistency.}
\label{tab:example-annotation}
\end{table}

\section{Introduction}
Question Generation (QG) is a typical Natural Language Generation (NLG) task that aims to generate natural language questions based on an input context and optionally an answer. QG has broad applications such as question answering (QA) \cite{lyu-etal-2021-improving}, conversational systems \cite{zeng-etal-2023-synthesize}, and knowledge assessment \cite{ghanem-etal-2022-question}. 
However, it has been demonstrated that questions generated by QG models suffer from problems like ambiguities and hallucinations \cite{laban-etal-2022-quiz}, which emphasizes the critical importance of reliable evaluations.

Human evaluation is widely acknowledged as the gold standard for evaluating QG \cite{wang-etal-2022-qrelscore}, with most automatic metrics striving to align their results with human evaluation results \cite{amidei-etal-2018-evaluation,10.1145/3485766}. However, the criteria of human evaluations are varied in existing research, leading to inconsistent and unreliable evaluations of QG models \cite{e24111514} and automatic metrics \cite{amidei-etal-2018-evaluation,mulla2023automatic}. This inconsistency highlights the urgent need to establish a unified human evaluation benchmark to ensure reliable evaluations.

Despite the importance of such benchmarks, few have been published, and the existing ones usually have the following limitations: 1) focusing only on specific dimensions like answerability; 2) involving a small amount of data (e.g., <1k samples); 3) employing a limited variety of models to generate questions, resulting in a lack of diversity in the data.  For instance, \citet{nema-khapra-2018-towards} generated monotonous questions using rule-based methods for evaluation and focused primarily on the answerability of questions, neglecting other dimensions.  \citet{gollapalli-ng-2022-qsts} evaluated generated questions from four dimensions but only included 500 questions generated by three models. \citet{laban-etal-2022-quiz} utilized seven QG models to generate 1k+ questions to be evaluated, however, they merely assessed whether the generated questions could be accepted as reading comprehension quiz questions, rather than scoring them on multiple dimensions. 

To address the above issues, we propose QGEval, a multi-dimensional evaluation benchmark, which evaluates questions across 7 dimensions and contains 3k questions generated by 15 QG models (including LLMs) based on 200 passages and answers. Specifically, through preliminary error analysis of the generated questions (described in section~\ref{sec:evaluation_methodology}), we identified seven evaluation dimensions and categorized them into two aspects: \textbf{1) Linguistic dimensions,} including fluency, clarity, and conciseness, which are basic requirements that a natural language text should meet; and \textbf{2) Task-oriented dimensions,} including relevance, consistency, answerability, and answer consistency, which involve requirements specific to QG tasks. 

As illustrated in Table~\ref{tab:example-annotation}, both linguistic and task-oriented dimensions are essential for a comprehensive evaluation of generated questions. In particular, although Q1 receives high scores in all task-oriented dimensions, it has a lower fluency score in linguistic dimensions due to the incorrect use of the interrogative word. On the contrary, Q2 performs well across all linguistic dimensions but scores poorly on most task-oriented dimensions because of its inconsistencies with the passage. These examples demonstrate the necessity of the two categories of evaluation dimensions.

Using QGEval to evaluate the performance of 15 different QG models, we find that these models perform relatively poorly in terms of answerability and answer consistency compared to other dimensions. We also evaluate and compare the performance of 15 existing automatic metrics, observing that there is still a gap between these metrics and human evaluations.

To summarize, our main contribution is four-fold: 
\begin{itemize}
    \item We introduce a multi-dimensional evaluation benchmark for QG named QGEval, which assesses the quality of questions across 7 dimensions and contains 3k questions generated by 15 QG models.
    \item We conduct a detailed analysis of the generated questions and compare the generation performance of various QG models across the seven dimensions, discovering that most models underperform in answerability and answer consistency.
    \item We evaluate and compare the performance of 15 automatic metrics across the seven dimensions, highlighting the discrepancies between automatic metrics and human evaluation.
    \item We have made the QGEval dataset, along with the codes for the automatic metrics we utilized, publicly accessible for further research.\footnote{Our data and code are publicly available at \url{https://github.com/WeipingFu/QGEval/}}
\end{itemize}

\begin{figure*}[t]
\includegraphics[width=\textwidth]{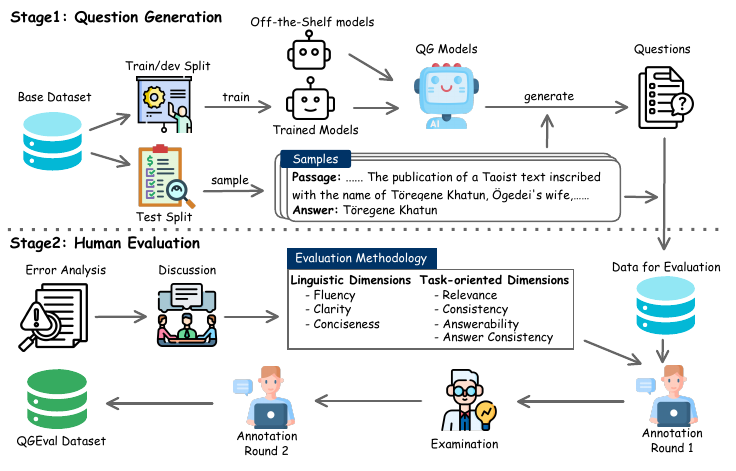}
\caption{Pipeline of dataset construction. Stage 1: Generate questions to be evaluated. Stage 2: Conduct two rounds of annotation to form the QGEval dataset.} 
\label{fig:pipeline}
\end{figure*}

\section{The QGEval Dataset}
In this section, we describe how we construct the QGEval dataset, the overall pipeline includes two stages: question generation and human evaluation, as shown in Figure~\ref{fig:pipeline}.

\subsection{Question Generation}
In the first stage, our goal is to generate questions for evaluation. We use SQuAD \cite{rajpurkar-etal-2016-squad} and HotpotQA \cite{yang-etal-2018-hotpotqa} as the base datasets, which are two widely used datasets in the field of QA and QG. We divide the SQuAD dataset into train/dev/test splits following \cite{10.1007/978-3-319-73618-1_56}. As for the HotpotQA dataset, we utilize its official train split and designate the first 3700 samples from the official dev set as our dev split and the rest as the test split. The train and dev splits are used to train QG models and the test split is then utilized to generate questions. We randomly select 100 samples from the test split of each dataset and utilize the passage and answer pairs provided by these samples to generate questions. The process results in the dataset to be evaluated, which comprises 3000 questions generated by multiple QG models based on 200 passages and answers. QG models contain both Off-the-Shelf models (public ones already trained on the QG task) and models trained by ourselves, the implementation details of QG models are in Appendix~\ref{apx:qgmodel_details}. 

To capture a wide diversity of model outputs and facilitate comparisons between different models and settings, our selection of QG models covers a variety of model sizes, types, and settings. Specifically, we utilize 14 QG models based on different language models and under various settings. The language models cover a broad range of sizes and encompass four different series of models: BART \cite{lewis-etal-2020-bart}, T5 \cite{2020t5}, Flan-T5 \cite{JMLR:v25:23-0870}, and GPT (OpenAI\footnote{https://platform.openai.com/docs/models}). Settings include fine-tuning, low-rank adaptation (LoRA) \cite{hu2022lora}, few-shot, and zero-shot. We customize the settings for models of different sizes, ensuring that each model is equipped with settings suitable for its characteristics. We also regard the references as outputs from one model for subsequent annotation, along with those from the other 14 models. Table~\ref{tab:models} shows all language model variants, the number of models' parameters, and the settings we employed for each model.

\begin{table}[t]
    \small
    \begin{tabularx}{\linewidth}{p{2.2cm}p{1.1cm}p{4cm}}
    \toprule
    \textbf{Model} & \textbf{Param.} & \textbf{Settings}\\
    \midrule
    BART-base & 140M & fine-tuning\\	 	 	 	 	 
    BART-large & 400M & fine-tuning\\
    \midrule
    T5-base & 250M & fine-tuning\\
    T5-large & 780M & fine-tuning\\
    \midrule
    Flan-T5-base & 250M & fine-tuning\\	 
    Flan-T5-large & 780M & fine-tuning\\
    Flan-T5-XL & 3B & \parbox{4cm}{LoRA;few-shot(8)}\\
    Flan-T5-XXL & 11B & \parbox{4cm}{LoRA;few-shot(8)}\\
    \midrule
    GPT-3.5-turbo & --- & \parbox{4cm}{few-shot(8);zero-shot}\\
    GPT-4 & --- & \parbox{4cm}{few-shot(8);zero-shot}\\
    \bottomrule
    \end{tabularx}
    \caption{Language models and settings used for question generation. GPT-4 refers to GPT-4-1106-preview. Since the parameter sizes of GPT-3.5-turbo and GPT-4-1160-preview have not been officially announced, we do not list them here.}
    \label{tab:models}
\end{table}

\subsection{Human Evaluation}
In the second stage, our objective is to obtain human ratings for each generated question. The evaluation methodology and the process of human annotation will be described in detail.

\paragraph{Evaluation Methodology}
\label{sec:evaluation_methodology}
To figure out which dimensions we should evaluate questions on, we conducted a pilot experiment to analyze the errors presented in the generated questions (see details in Appendix~\ref{apx:error_analysis}). We observed that QG models may generate questions that are incorrectly formed (e.g., not a question) and phrased, ambiguous, or verbose, making it difficult to understand their intent. QG models may also generate questions that are irrelevant to the context, inconsistent with the provided information, unanswerable, or mismatched with the given answer, failing to meet the requirements of the QG task. From these observations, we conclude that the errors can be categorized into two types: linguistic and task-oriented. After a thorough discussion with two experts in the field of education, we determined that the quality of questions should be evaluated on the following seven dimensions, including both linguistic and task-oriented aspects.

Linguistic dimensions serve as the foundational evaluation dimensions in most NLG tasks including QG. Specifically, we focus on the following three linguistic dimensions in our evaluation, requiring the generated questions to be well-formed, and expressed clearly and concisely. 
\begin{itemize}
    \item \textbf{Fluency (Flu.):} Whether the question is well-formed, grammatically correct, coherent, and fluent enough to be understood \cite{oh-etal-2023-evaluation}.
    \item \textbf{Clarity (Clar.):} Whether the question is expressed clearly and unambiguously, avoiding excessive generality and ambiguity, the same as the definition in \cite{ousidhoum-etal-2022-varifocal}.
    \item \textbf{Conciseness (Conc.):} Whether the question is concise and not abnormally verbose with redundant modifiers, as defined in \cite{cheng-etal-2021-guiding}.
\end{itemize}

Task-oriented dimensions refer to those aspects associated with the QG task, measuring the correlation between the generated questions and passages, as well as the connection between questions and the provided answers. The task-oriented dimensions we considered are outlined below, requiring the generated questions to be contextually relevant and consistent, answerable based on the passage, and match the provided answers.
\begin{itemize}
    \item \textbf{Relevance (Rel.):} Whether the question is relevant to the given passage and asks for key information from the passage. It is also a commonly used dimension in both QG and other text generation tasks \cite{oh-etal-2023-evaluation,10.1145/3485766}.
    \item \textbf{Consistency (Cons.):} Whether the information presented in the question is consistent with the passage and without any contradictions or hallucinations, similar to the definition in other text generation tasks \cite{honovich-etal-2022-true-evaluating}.
    \item \textbf{Answerability (Ans.):} Whether the question can be distinctly answered based on the passage, a widely used and distinctive dimension in QG \cite{ghanem-etal-2022-question}. 
    \item \textbf{Answer Consistency (AnsC.):} Whether the question can be answered using the provided answer, as "Answer Matching" defined in \cite{cheng-etal-2021-guiding}.
\end{itemize}

The scoring scale for each dimension is 1 to 3, with higher being better (detailed scoring guidelines are presented in the Appendix~\ref{apx:annotation_instruct}).

\begin{table*}[t]
\centering
\small
    \begin{tabularx}{\textwidth}{X@{\hspace{2em}}XXXXXXX}
    \toprule
    \textbf{Rounds} & \textbf{Flu.} & \textbf{Clar.} & \textbf{Conc.} & \textbf{Rel.} & \textbf{Cons.} & \textbf{Ans.} & \textbf{AnsC.}\\
    \midrule
    Round1 & 0.226 & 0.375 & 0.515 & 0.233 & 0.181 & 0.354 & 0.559\\	 
    Round2 & 0.427 & 0.576 & 0.755 & 0.437 & 0.445 & 0.661 & 0.800\\	
    \bottomrule
    \end{tabularx}
    \caption{Krippendorff's alpha coefficient of inter-annotator scores in the first and second round annotations. A higher score means higher agreement among the annotators.}
    \label{tab:annotation_coeff}
\end{table*}
\paragraph{Annotation Process}
Due to the subjective nature of annotation, a crowdsourcing annotation approach was adopted. Three postgraduate students specializing in computer science volunteered as annotators to score the generated questions according to the detailed scoring guidelines for each dimension on our annotation platform (see the interface in Appendix~\ref{apx:annotation_interface}). The annotators are all specializing in the research fields of QG and QA and are all proficient in reading and writing in English. They are familiar with the task and have a good understanding of the annotation guidelines. Before the formal annotation process, a trial annotation involving 100 samples was conducted, and the results were reviewed by two educational experts. The trial results indicated that the three annotators were well-equipped to handle this task. 

Two rounds of annotation were performed in the formal annotation process to confirm judgments and ensure a higher quality of annotation results. In the first round, questions generated based on SQuAD and HotpotQA were presented and scored separately. For the same passage, all 15 questions generated by different models were presented simultaneously, and annotators scored these questions sequentially. Annotating in this way increases efficiency and helps annotators validate their judgments (e.g., similar questions should receive similar scores). During annotation, the generative models were kept unaware. Annotation results from the first round were examined. In the second round, the annotators were required to review samples that may have been incorrectly scored. For each dimension, the annotators: 1) checked annotations when the same questions received different scores on the same dimension; 2) reviewed samples where their annotations differed from the other annotators by 2 points, while the annotations of the other two annotators were the same; 3) discussed with each other when the annotation scores in the first round were 1, 2, 3. 

To assess the agreement between annotators, Krippendorff's alpha coefficient, a statistical measure of inter-rater reliability, was calculated for each dimension and shown in Table~\ref{tab:annotation_coeff}. In the first round, the coefficients ranged from 0.181 to 0.559 and improved to a range of 0.427 to 0.800 in the second round. Furthermore, to verify the quality of annotations, 100 samples were randomly selected and reviewed by the two experts. The results showed that the accuracy of annotations for each dimension was over 96\%. 

\section{Experiment and Evaluation}
In this section, we conduct a series of analytical experiments and evaluations on QG models and automatic metrics with QGEval. We aim to address the following three research questions.

\begin{figure*}[t]
    \includegraphics[width=\textwidth]{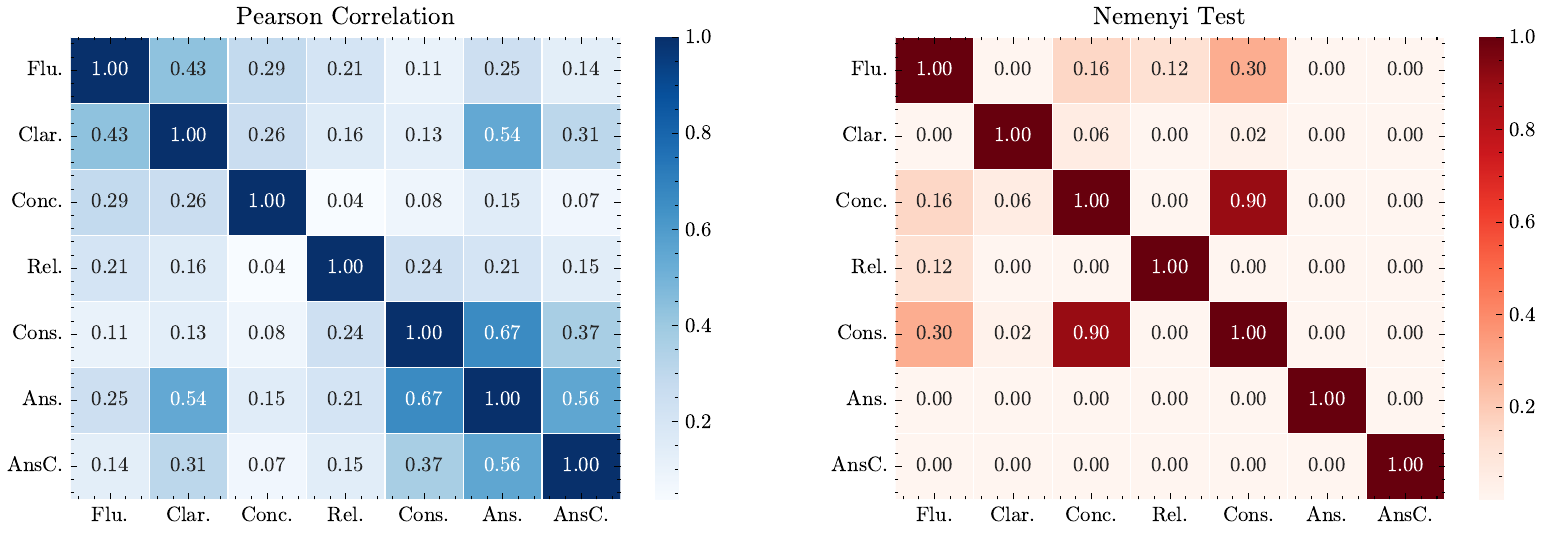}
    \caption{Pearson correlations and p-values of Nemenyi test for seven dimensions.}
    \label{fig:dimension_heatmap}
\end{figure*}

\subsection{Are the seven dimensions appropriate for the evaluation of QG?}
To figure out whether the dimensions are an appropriate set, we examine the correlations and distinctions among them by calculating Pearson correlations and conducting the Nemenyi test on the annotation scores for these dimensions. Pearson correlation measures the linear correlation between two sets of data, with higher absolute values indicating stronger correlations. The Nemenyi test determines whether there are significant differences between groups, with lower p-values indicating greater significance. Intuitively, the seven dimensions might correlate with each other but should also maintain differences from one another. As shown in Figure~\ref{fig:dimension_heatmap}, the Pearson correlations between the seven dimensions are within a reasonable range (0.04 to 0.67), and most p-values in the Nemenyi test are below 0.05. This indicates that \textbf{these dimensions are interrelated but still exhibit distinct characteristics}, consistent with our intuition. 

For correlations between dimensions, we further observe: 
1) The correlation coefficients among the linguistic dimensions (fluency, clarity, and conciseness) are relatively high.
2) Linguistic dimensions can influence task-oriented dimensions. For instance, clarity and consistency show high correlations with answerability. Unclear expression (low clarity) and contradictions between the question and passage (low consistency) may lead to a low score of answerability.
3) As expected, answer consistency is highly relevant to answerability, and from experience, unanswerable questions tend to have low answer consistency scores.

\subsection{How do the QG models perform across the seven dimensions?}
\label{subsec:performance_of_model}
By asking this question, we aim to explore which dimensions QG models perform well or poorly on and to compare the generation performance of different QG models. Table~\ref{tab:annotation_result} shows the averaged annotation scores along seven evaluation dimensions of all QG models. Generally speaking, \textbf{most QG models are capable of generating questions that are both fluent and relevant to the provided passage}, i.e., received high ratings on both fluency and relevance dimensions. However, they often \textbf{encounter challenges in generating questions that are answerable and align well with the given answers.} Inspired by this finding, we advocate that future question generation work should focus more on improving the answerability and answer consistency of generated questions.

\begin{figure*}[t]
    \includegraphics[width=\textwidth]{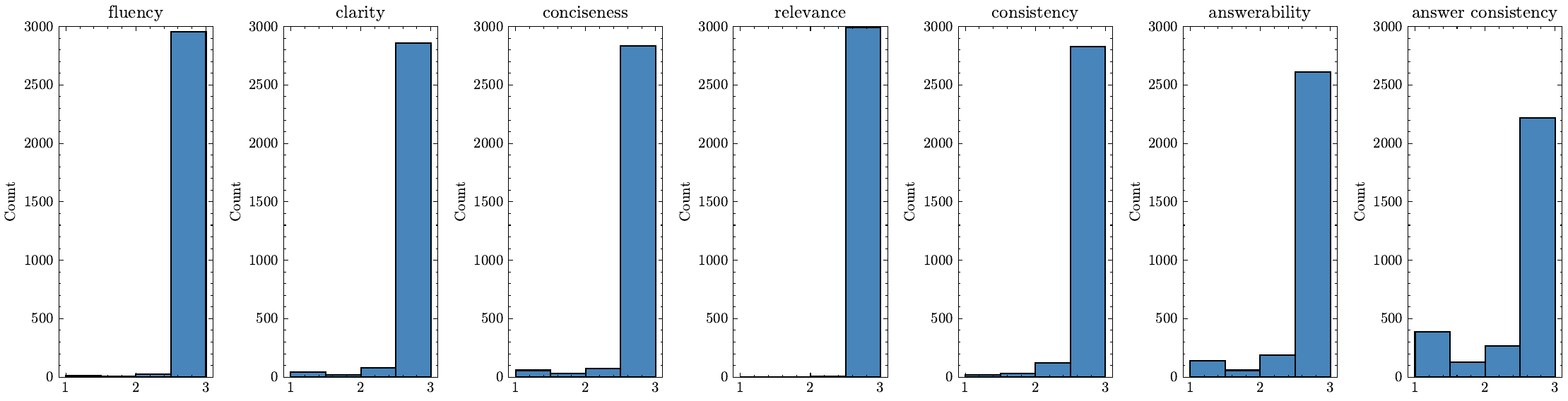}
    \caption{Annotation score distributions across seven dimensions.}
    \label{fig:annotation_dist_all}
\end{figure*}

We also observe that the average scores of these models are high (above 2). We further take a look into the annotation score distribution in Figure~\ref{fig:annotation_dist_all} and find that most labels are rated 3, with 1 and 2 being rare, which indicates that the proportion of poorly performed questions among the generated questions over the 7 dimensions is small (particularly in fluency and relevance). Rating 3 accounts for a relatively small proportion of answerability and answer consistency, which also suggests that the generated questions are deficient in the two dimensions.

\begin{table*}[t]
\centering
\small
    \begin{tabularx}{\textwidth}{lXXXXXXX>{\columncolor{blue!10}}X}
    \toprule
        \textbf{Models} & \textbf{Flu.} & \textbf{Clar.} & \textbf{Conc.} & \textbf{Rel.} & \textbf{Cons.} & \textbf{Ans.} & \textbf{AnsC.} & \textbf{Avg.}\\
    \midrule
        Reference & 2.968 & 2.930 & \textbf{2.998} & 2.993 & 2.923 & 2.832 & \textbf{2.768} & \textbf{2.916}\\
        BART-base-finetune & 2.958 & 2.882 & 2.898 & 2.995 & 2.920 & \underline{2.732} & 2.588 & 2.853\\
        BART-large-finetune & \underline{2.932} & 2.915 & \underline{2.828} & 2.995 & 2.935 & 2.825 & \textbf{2.737} & 2.881\\
        T5-base-finetune & 2.972 & 2.923 & 2.922 & \textbf{3.000} & \underline{2.917} & 2.788 & 2.652 & 2.882\\ 
        T5-large-finetune & 2.978 & 2.930 & 2.907 & 2.995 & 2.933 & 2.795 & 2.720 & 2.894\\
        Flan-T5-base-finetune & 2.963 & 2.888 & 2.938 & \textbf{2.998} & 2.925 & 2.775 & 2.665 & 2.879\\
        Flan-T5-large-finetune & 2.982 & 2.902 & 2.895 & 2.995 & \textbf{2.950} & 2.818 & 2.727 & 2.895\\
        Flan-T5-XL-LoRA & \underline{2.913} & \underline{2.843} & \underline{2.880} & 2.997 & 2.928 & 2.772 & 2.667 & 2.857\\
        Flan-T5-XXL-LoRA & \underline{2.938} & \underline{2.848} & 2.907 & \textbf{3.000} & 2.943 & 2.757 & 2.678 & 2.867\\
        Flan-T5-XL-fewshot & 2.975 & \underline{2.820} & \textbf{2.985} & \underline{2.955} & \underline{2.908} & \underline{2.652} & \underline{2.193} & \underline{2.784}\\ 
        Flan-T5-XXL-fewshot & \textbf{2.987} & 2.882 & \textbf{2.990} & \underline{2.988} & 2.920 & \underline{2.687} & 2.432 & \underline{2.841}\\
        GPT-3.5-turbo-fewshot & 2.972 & 2.927 & \underline{2.858} & 2.995 & \textbf{2.955} & \textbf{2.850} & \underline{2.335} & 2.842\\
        GPT-4-fewshot & \textbf{2.988} & \textbf{2.987} & 2.897 & 2.992 & \textbf{2.947} & \textbf{2.922} & \textbf{2.772} & \textbf{2.929}\\
        GPT-3.5-turbo-zeroshot & \textbf{2.995} & \textbf{2.977} & 2.913 & 2.992 & \underline{2.917} & 2.823 & \underline{2.157} & \underline{2.825}\\
        GPT-4-zeroshot & 2.983 & \textbf{2.990} & 2.943 & \underline{2.970} & 2.932 & \textbf{2.883} & 2.723 & \textbf{2.918}\\
    \midrule
        Avg. & 2.967 & 2.910 & 2.917 & 2.991 & 2.930 & 2.794 & 2.588 \\
    \bottomrule
    \end{tabularx}
\caption{Annotation scores of questions along seven dimensions, averaged over three annotators. The three highest and lowest scores of each dimension are bolded and underlined, respectively. GPT-4 refers to GPT-4-1106-preview. Avg. refers to the average score.}
\label{tab:annotation_result}
\end{table*}

\textbf{We further compare these models via different model sizes and settings}, our findings are:
1) The best three QG models ranked by the average scores of all dimensions are GPT-4-fewshot, GPT-4-zeroshot, and reference, indicating that the quality of questions generated by GPT-4 is comparable to that of humans.
2) Under the same setting, as the model size increases, the generated questions exhibit improved clarity in expression, higher consistency with the provided passages, and increased alignment with the provided answers.
3) Maintaining the same model, the zero-shot approach performs less effectively than the few-shot approach, and the few-shot approach is inferior to the supervised (LoRA) approach, especially on the consistency, answerability, and answer consistency dimensions.
4) Models under zero-shot and few-shot settings often fail to generate questions that match the given answers, except for GPT-4, which could be due to the models' insufficient ability to follow detailed instructions. 

To assess the benchmark's discriminative power among different models, we conducted t-tests comparing the scores of models ranked in various percentiles: 1 (top 6\%) vs. -1 (bottom 6\%), 3 (top 20\%) vs. -3 (bottom 20\%), and 5 (top 33\%) vs. -5 (bottom 33\%) across each dimension (detailed results are presented in Appendix~\ref{apx:discriminative_power}). \textbf{The results indicate that the benchmark demonstrates limited discriminative power.} Except for answer consistency, the top-five performing models fail to exhibit significant differences compared to the bottom-five models across the other six dimensions. For instance, in fluency, the t-test p-value between GPT-3.5-turbo-zeroshot (top 1) and Flan-T5-XLLoRA (bottom 1) is below 0.05, suggesting a significant difference. Conversely, the t-test p-value between Flan-T5-large-finetune (top 5) and FlanT5-base-finetune (bottom 5) is much higher than
0.05, indicating only a minor difference. Although these dimensions do not show strong discriminative power among current QG models, they are still frequently used in recent research. We advocate exploring more discriminative and advanced dimensions beyond the basic ones, such as whether the question involves key content, the novelty of the question, its ability to guide deeper thinking, etc. We believe that generated questions should meet the requirements of basic dimensions explored in our work before they can satisfy such advanced dimensions.

\subsection{Can existing automatic metrics accurately evaluate generated questions?}
In this section, we use QGEval to evaluate and compare the performance of existing automatic metrics to find out whether these metrics are able to accurately evaluate the quality of generated questions across the seven dimensions.

\paragraph{Automatic Metric}
Our selection of automatic metrics varies from methods based on lexical overlap to those based on large language models, including both reference-based and reference-free approaches.  Reference-based metrics evaluate questions by computing the similarity between them and the references, which include BLEU \cite{papineni2002bleu}, ROUGE \cite{lin-2004-rouge}, METEOR \cite{banarjee2005}, BERTScore \cite{Zhang*2020BERTScore:}, MoverScore \cite{zhao-etal-2019-moverscore}, BLEURT \cite{sellam-etal-2020-bleurt}, Q-Metric \cite{nema-khapra-2018-towards}, and QSTS \cite{gollapalli-ng-2022-qsts}. Reference-free metrics utilize the comprehension and generation capabilities of language models to evaluate questions without references, including BARTScore \cite{NEURIPS2021_e4d2b6e6}, GPTScore \cite{fu2023gptscore}, UniEval \cite{zhong-etal-2022-towards}, QRelScore \cite{wang-etal-2022-qrelscore}, and RQUGE \cite{mohammadshahi-etal-2023-rquge}. When using the ref-hypo scoring type (generate candidate text based on the reference), BARTScore and GPTScore are considered reference-based metrics. Among all these metrics, UniEval and GPTScore are designed for multi-dimensional evaluation, offering a score for each dimension (7 scores for 7 dimensions), while the other metrics provide only a single overall score.
Detailed descriptions of these metrics are presented in Appendix~\ref{apx:metrics}.

\paragraph{Metric Evaluation}
We evaluate the agreement between the automatic metrics and human annotation scores by calculating the Pearson correlation over each dimension, results are shown in Table~\ref{tab:metric_annotation_corr}, with the three highest and lowest absolute coefficients bolded and underlined respectively. 
\begin{table*}[t]
    \centering
    \small
    \begin{tabularx}{\textwidth}{p{2.5cm}XXXXXXX}
        \toprule
            \textbf{Metrics} & \textbf{Flu.} & \textbf{Clar.} & \textbf{Conc.} & \textbf{Rel.} & \textbf{Cons.} & \textbf{Ans.} & \textbf{AnsC.}\\
        \midrule
            \multicolumn{8}{c}{\textit{Reference-based Metrics}} \\
        \midrule
            BLEU-4 & \underline{0.028} & \underline{0.049} & 0.138 & \underline{0.041} & \underline{0.032} & \underline{0.080} & \underline{0.162}\\	
            ROUGE-L & 0.080 & 0.086 & 0.234 & 0.085 & 0.079 & 0.127 & 0.233 \\
            METEOR & \underline{0.020} & 0.088 & \underline{0.106} & 0.079 & 0.059 & 0.131 & 0.253\\	 	 		 	 	 	 
            BERTScore & 0.140 & \textbf{0.123} & \textbf{0.313} & \textbf{0.113} & 0.091 & 0.131 & 0.231\\ 	 	 
            MoverScore & 0.070 & \underline{0.075} & 0.209 & 0.071 & 0.058 & 0.101 & 0.188 \\
            BLEURT & 0.078 & \textbf{0.105} & 0.179 & 0.104 & 0.098 & 0.144 & \textbf{0.271}\\
            BARTScore-ref & 0.087 & 0.079 & 0.235 & 0.109 & 0.078 & 0.092 & 0.190\\
            GPTScore-ref & 0.069 & 0.086 & 0.182 & \underline{0.006} & 0.054 & 0.106 & 0.187\\
            Q-BLEU4	& 0.072 & 0.082 & 0.216 & 0.058 & 0.075 & 0.113 & 0.198\\
            QSTS &\underline{0.016} & 0.104 & \underline{0.015} & 0.077 & 0.043 & 0.130 & 0.250\\
        \midrule
            \multicolumn{8}{c}{\textit{Reference-free Metrics}} \\
        \midrule
            BARTScore-src & \textbf{-0.148} & \underline{-0.035} & \textbf{-0.511} & 0.053 & \underline{-0.001} & \underline{0.018} & \underline{-0.015}\\
            GPTScore-src & 0.134 & 0.104 & \underline{-0.052} & \textbf{0.416} & \textbf{0.197} & \textbf{0.148} & 0.236\\
            UniEval & \textbf{0.370} & \textbf{0.219} & 0.259 & \textbf{0.153} & \textbf{0.156} & \textbf{0.207} & \textbf{0.356}\\
            QRelScore & \textbf{-0.213} & -0.096 & \textbf{-0.553} & \underline{0.032} & \underline{0.002} & \underline{-0.026} & \underline{-0.025}\\
            RQUGE & 0.045 & 0.092 & 0.126 & 0.070 & \textbf{0.200} & \textbf{0.211} & \textbf{0.561}\\
        \bottomrule
    \end{tabularx}
    \caption{Pearson correlation between automatic metrics and human scores along seven dimensions. The three highest and lowest absolute coefficients of each dimension are bolded and underlined, respectively. BLEU-4: 4-gram variant of BLEU; ROUGE-L: the longest common subsequence (LCS) variant of ROUGE; *-ref: ref-hypo scoring type, *-src: src-hypo scoring type.}
    \label{tab:metric_annotation_corr}
\end{table*}

\begin{table}[t]
    \centering
    \small
    \begin{tabularx}{\linewidth}{p{1.4cm}p{1.2cm}|p{2.4cm}p{1cm}}
    \toprule
        \textbf{Metrics} & \textbf{Pearson} & \textbf{Metrics} & \textbf{Pearson}\\
    \midrule
        GPTScore & 0.187 & GPT-3.5 & 0.195\\
        UniEval & 0.215 & G-EVAL\textsubscript{GPT-3.5} & 0.228\\
        RQUGE & 0.250 & \textbf{GPT-4} & \textbf{0.296}\\
         & & \textbf{G-EVAL\textsubscript{GPT-4}} & \textbf{0.356}\\
    \bottomrule
    \end{tabularx}
    \caption{Pearson correlation between annotation scores and metrics on answerability. GPT-3.5 and GPT-4 refer to the methods using direct prompts. GPTScore refers to GPTScore-src.}
    \label{tab:LLM_eval_result}
\end{table}

Correlation results show several trends. 
1) \textbf{Most metrics have relatively low correlations with annotation scores along seven dimensions,} ranging from -0.4 to 0.4, especially on fluency, clarity, relevance, and consistency. We observed that most questions received high annotation scores across these four dimensions, while the scores assigned by automatic metrics varied significantly, resulting in poor alignment with human scores.
2) In general, \textbf{reference-free metrics tend to outperform reference-based metrics,} exhibiting higher correlation coefficients with human evaluation. 
3) \textbf{Metrics that conduct multi-dimensional evaluations tend to perform better across a wider range of dimensions} compared to those that provide only a single composite score (BLEU, ROUGE, BARTScore, etc.). UniEval, for example, achieves the three highest coefficients across six dimensions.
4) \textbf{Metrics designed for specific dimensions are better than other metrics on those specific dimensions.} RQUGE, leveraging question-answering results for evaluation, attains higher correlations on its target dimensions: answerability and answer consistency. 
The observations in 3) and 4) imply that metrics with a single composite score are not suitable for the comprehensive evaluation of generated questions. Instead, \textbf{designing multi-dimensional metrics or metrics focused on specific dimensions may yield better results.}

Our further exploration of the score distribution of automatic metrics (in Appendix~\ref{apx:dis_metric}) and the application of these metrics to rank different QG models (in Appendix~\ref{apx:metric_rank_model}) indicates that existing automatic metrics still struggle to effectively distinguish questions of varying quality.

\paragraph{LLM as Evaluator}
Recent work has leveraged LLMs for NLG evaluation and found that LLM-based metrics are superior to former metrics \cite{kocmi-federmann-2023-large}. To assess the effectiveness of employing LLMs for question generation evaluation, we use the GPT-3.5-turbo and GPT-4-1106-preview as evaluators and implement evaluations using both direct prompts and G-EVAL \cite{liu-etal-2023-g}, an evaluation method employing Chain-of-Thought (COT). Due to budget constraints, we conducted tests on 450 questions (30 passages) solely focusing on the answerability dimension for analysis. The Pearson correlations between annotation scores and metrics are shown in Table~\ref{tab:LLM_eval_result}.

We compare the performance of LLM-based metrics with RQUGE, UniEval, and GPTScore-src (the top three metrics on answerability in Table~\ref{tab:metric_annotation_corr}) here. The results show that metrics based on GPT-4 achieve the highest correlations with human scores, which demonstrates the potential of using LLMs for QG evaluation. The comparisons between methods using direct prompts and G-EVAL also verify the effectiveness of COT. Although LLM-based metrics outperform other evaluation methods, they still fail to align closely with human evaluation (Pearson correlations are below 0.4). Further exploration is needed in future work.  

\section{Related Work}
\paragraph{Automatic Metrics}
Automatic evaluation of QG is still dominated by reference-based metrics such as BLEU \cite{papineni2002bleu} and Q-Metric \cite{nema-khapra-2018-towards}, which compute the similarity between generated questions and references. As QG is a one-to-many generation task, this type of metric can not evaluate questions that are different from the references \cite{mohammadshahi-etal-2023-rquge}. Reference-free metrics like BARTScore \cite{NEURIPS2021_e4d2b6e6} and QRelScore \cite{wang-etal-2022-qrelscore} overcome this limitation, but they often assign a single overall score as the evaluation result, which is less interpretable and not comprehensive. UniEval \cite{zhong-etal-2022-towards} and GPTScore \cite{fu2023gptscore} are designed to evaluate generated texts from multiple interpretable dimensions, but they are not specifically designed for the QG task, and thus their performance in evaluating QG is limited.

\paragraph{Human Evaluation in QG}
Since existing automatic metrics are not effective enough to measure the quality of generated questions, human evaluation is frequently used in the field of QG \cite{mulla2023automatic}. However, the human evaluation criteria provided by existing works are disparate, leading to inconsistent evaluation of generated questions.
\citet{ghanem-etal-2022-question} utilized answerability, fluency, and grammaticality to assess question quality, while \citet{ushio-etal-2022-generative} employed grammatically, understandability, and answerability for evaluation. \citet{gou-etal-2023-diversify} focused on consistency and diversity of generated questions. Disparate human evaluation criteria also result in inconsistent evaluation and comparison of automatic metrics. \citet{nema-khapra-2018-towards} proposed Q-metric and computed the correlations between existing automatic metrics and human judgments on answerability. \citet{wang-etal-2022-qrelscore} proposed QRelScore and compared it with other metrics based on their human evaluation results on three dimensions: grammaticality, relevance, and answerability. \citet{mohammadshahi-etal-2023-rquge} evaluated the performance of automatic metrics on their newly annotated data as the human evaluation of generated questions is not available in previous work. Thus, it's urgent to develop unified and reliable human evaluation benchmarks to ensure consistent and accurate assessments of generated questions and automatic metrics.

\section{Conclusion}
In this work, we introduced a comprehensive, multi-dimensional evaluation benchmark, QGEval, to facilitate the evaluation of generated questions from various models and existing automatic metrics across 7 dimensions: fluency, clarity, conciseness, relevance, consistency, answerability, and answer consistency. It contains 3k questions generated from 15 different QG models.
Through analysis of QGEval, we found that most models performed unsatisfactorily on answerability and answer consistency. This highlights the importance of focusing on the two dimensions in future QG model designs. 
Additionally, our evaluation of 15 existing automatic metrics revealed that these metrics still exhibit relatively low correlation coefficients with human annotation scores, emphasizing the need to explore advanced metrics that align better with human evaluation. We hope that this work will serve as a valuable resource for future research on question generation evaluation and models.

\section{Limitations}
Our work proposes QGEval, a multi-dimensional evaluation benchmark for QG, to evaluate and compare the performance of different QG models and existing automatic metrics. Although it provides a comprehensive evaluation of generated questions, it still has the following two limitations.

First, it focuses on the scenario of generating questions based on a passage and an optional answer and is not applicable to other scenarios such as visual question generation \cite{vedd-etal-2022-guiding} and conversational question generation \cite{zeng-etal-2023-synthesize}. Additional dimensions may be introduced to meet some specific requirements. For example, complexity is considered when the generated questions are required to involve multi-hop reasoning \cite{fei-etal-2022-cqg}. In this work, we consider more general requirements under the scenario we focus on.

Second, the proposed dimensions have limited discriminative power for current QG models based on pre-trained language models (as discussed in~\ref{subsec:performance_of_model}). Most of these QG models perform well across the seven dimensions, particularly in fluency and relevance (questions rated 3 account for a large proportion in the two dimensions). Except for the seven basic dimensions we explored in our work, we advocate the exploration of more discriminative and advanced dimensions, such as the inclusion of key content, the novelty of the question, its potential to foster critical thinking and deeper engagement, etc.

\section{Acknowledgments}
This work was supported by National Key Research and Development Program of China (2022YFC3303600), National Natural Science Foundation of China (62137002, 62293553, 62293554, 62437002, and 62176209), "LENOVO-XJTU" Intelligent Industry Joint Laboratory Project, Natural Science Basic Research Program of Shaanxi (2023-JC-YB-593), the Youth Innovation Team of Shaanxi Universities, Project of China Knowledge Centre for Engineering Science and Technology.

\bibliography{custom}

\clearpage
\appendix
\section{Annotation Details}
\subsection{Annotation Instructions and Examples}
\label{apx:annotation_instruct}
The generated questions are rated on a scale of 1 to 3 for each dimension. detailed scoring guidelines are shown in Table~\ref{tab:annotation_ins}. Table~\ref{tab:annotation_example} also provides several annotation examples. The first three examples (Example 1 to Example 3) present questions that highlight issues related to linguistic dimensions. From these examples, we observe that fluency can affect clarity, and conciseness has a certain impact on fluency. Additionally, these examples demonstrate how linguistic dimensions influence task-oriented dimensions; for instance, answerability can be significantly influenced by fluency and clarity, while conciseness has a comparatively minor effect. Example 4 and Example 5 illustrate that low consistency and low answerability can lead to low answer consistency. Conversely, answer consistency can also receive a low rating even when both consistency and answerability are rated high. Example 6 presents a good question that received high scores across all seven dimensions. 
\begin{table*}[htbp]
    \centering
    \small
    \begin{tabularx}{\textwidth}{Xl}
        \toprule
        \textbf{Dimensions} & \textbf{Instructions} \\
        \midrule
        \textbf{Fluency} & 
        \parbox{13cm}{
            Score 1: The question is incoherent, with imprecise wording or significant grammatical errors, making it difficult to comprehend its meaning.\\
            Score 2: The question is slightly incoherent or contains minor grammatical errors, but it does not hinder the understanding of the question's meaning.\\
            Score 3: The question is fluent and grammatically correct.}\\
        \midrule
        \textbf{Clarity} & 
        \parbox{13cm}{
            Score 1: The question is too broad or expressed in a confusing manner, making it difficult to understand or leading to ambiguity. \textit{Particularly, if the generated sentence is not a question but a declarative sentence, it should be considered in this situation.}\\
            Score 2: The question is not expressed very clearly and specifically, but it is possible to infer the question's meaning based on the given passage.\\
            Score 3: The question is clear and specific, without any ambiguity.}\\
        \midrule
        \textbf{Conciseness} & 
        \parbox{13cm}{
            Score 1: The question contains too much redundant information, making it difficult to understand its intent.\\
            Score 2: The question includes some redundant information, but it does not impact the understanding of its meaning.\\
            Score 3: The question is concise and does not contain any unnecessary information.}\\
        \midrule
        \textbf{Relevance} & 
        \parbox{13cm}{
            Score 1: The question is completely unrelated to the passage.\\
            Score 2: The question is somewhat related to the passage and it asks for non-crucial information related to the passage.\\
            Score 3: The question is relevant to the context, and the information it seeks is crucial to the passage.}\\
        \midrule
        \textbf{Consistency} & 
        \parbox{13cm}{
            Score 1: The question contains factual contradictions with the passage or logical errors.\\
            Score 2: The information sought in the question is not fully described in the passage.\\
            Score 3: The information in the question is entirely consistent with the passage.}\\
        \midrule
        \textbf{Answerability} & 
        \parbox{13cm}{
            Score 1: The question cannot be answered based on the provided passage.\\
            Score 2: The question can be partially answered based on the provided passage, or the answer to the question can be inferred to some extent.\\
            Score 3: The question can be answered definitively based on the given passage.}\\
        \midrule
        \textbf{\parbox{2cm}{Answer \\ Consistency}} & 
        \parbox{13cm}{
            Score 1: The question cannot be answered by the provided answer.\\
            Score 2: The question can be partially answered using the provided answer.\\
            Score 3: The question can be answered directly using the provided answer.}\\
        \bottomrule
    \end{tabularx}
    \caption{Annotation instructions of evaluation dimensions.}
    \label{tab:annotation_ins}
\end{table*}

\subsection{Annotation Interface}
\label{apx:annotation_interface}
\begin{figure*}[htbp]
\includegraphics[width=\textwidth]{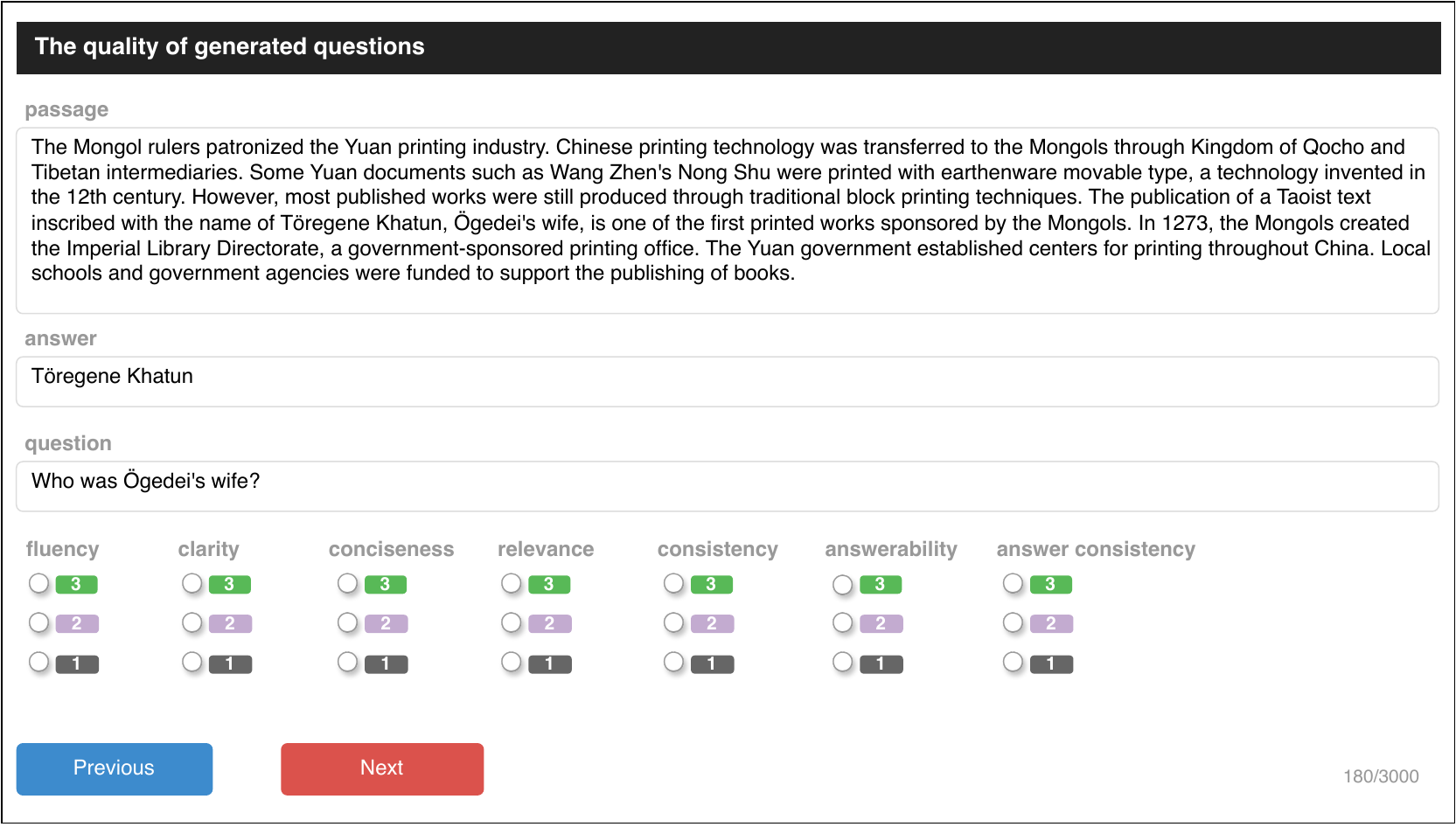}
\caption{Annotation interface.}
\label{fig:annotation_interface}
\end{figure*}

The annotation interface is presented in Figure~\ref{fig:annotation_interface}. In the annotation process, annotators should first carefully read the content of the given passage, answer, and question, and then select a score for each dimension.

\section{Implementation Details of QG models}
\label{apx:qgmodel_details}
QG models based on open-source language models are implemented using Hugging Face Transformers, while QG models based on closed-source language models utilize the official open API provided by the respective model. Detailed task instructions we applied for each QG model are presented in Table~\ref{tab:qgmodel_ins}.

\begin{table*}[htbp]
    \centering
    \small
    \begin{tabularx}{\textwidth}{X}
        \toprule
        \parbox{15.5cm}{
            \textbf{Example 1}\\
            \textbf{Passage:} Richard "Rick" Ducommun (July 3, 1952 – June 12, 2015) was a Canadian actor, comedian and writer who appeared in films and television. \\The 'Burbs is a 1989 American comedy thriller film directed by Joe Dante starring Tom Hanks, Bruce Dern, Carrie Fisher, Rick Ducommun, Corey Feldman, Wendy Schaal and Henry Gibson.  The film was written by Dana Olsen, who also has a cameo in the movie.  The film pokes fun at suburban environments and their eccentric dwellers.\\
            \textbf{Answer:} Richard "Rick" Ducommun\\
            \textbf{Question:} What star if the Burbs was Canadian?\\
            \textbf{Scores:} fluency - 1; clarity - 1; conciseness - 3; relevance - 3; consistency - 3; answerability - 1.3333; answer consistency - 1.3333
        }
        \tabularnewline
        \midrule
        \parbox{15.5cm}{
            \textbf{Example 2}\\
            \textbf{Passage:} At the same time the Mongols imported Central Asian Muslims to serve as administrators in China, the Mongols also sent Han Chinese and Khitans from China to serve as administrators over the Muslim population in Bukhara in Central Asia, using foreigners to curtail the power of the local peoples of both lands. Han Chinese were moved to Central Asian areas like Besh Baliq, Almaliq, and Samarqand by the Mongols where they worked as artisans and farmers. Alans were recruited into the Mongol forces with one unit called "Right Alan Guard" which was combined with "recently surrendered" soldiers, Mongols, and Chinese soldiers stationed in the area of the former Kingdom of Qocho and in Besh Balikh the Mongols established a Chinese military colony led by Chinese general Qi Kongzhi (Ch'i Kung-chih). After the Mongol conquest of Central Asia by Genghis Khan, foreigners were chosen as administrators and co-management with Chinese and Qara-Khitays (Khitans) of gardens and fields in Samarqand was put upon the Muslims as a requirement since Muslims were not allowed to manage without them. The Mongol appointed Governor of Samarqand was a Qara-Khitay (Khitan), held the title Taishi, familiar with Chinese culture his name was Ahai.\\
            \textbf{Answer:} artisans and farmers\\
            \textbf{Question:} Where did the Mongols work?\\
            \textbf{Scores:} fluency - 3; clarity - 1.6667; conciseness - 3; relevance - 3; consistency - 3; answerability - 1.3333; answer consistency - 1
        }
        \tabularnewline
        \midrule
        \parbox{15.5cm}{
            \textbf{Example 3}\\
            \textbf{Passage:} "Domino Dancing" is a song recorded by the British synthpop duo Pet Shop Boys, released as the lead single from their 1988 album, "Introspective".  It reached number 7 on the UK Singles Chart.\\Introspective is the third studio album by English synthpop duo Pet Shop Boys.  It was first released on 11 October 1988 and is the Pet Shop Boys' second-best-selling album, selling over 4.5 million copies worldwide.  (Their fifth studio album, "Very", sold more than 5 million copies worldwide.).\\
            \textbf{Answer:} October\\
            \textbf{Question:} "Domino Dancing" is a song recorded by the British synthpop duo Pet Shop Boys, released as the lead single from their 1988 album, "Introspective".  It reached number 7 on the UK Singles Chart, which month was the album "Introspective" first released?\\
            \textbf{Scores:} fluency - 2.3333; clarity - 3; conciseness - 1.6667; relevance - 3; consistency - 3; answerability - 3; answer consistency - 3
        }
        \tabularnewline
        \midrule
        \parbox{15.5cm}{
            \textbf{Example 4}\\
            \textbf{Passage:} With International Criminal Court trial dates in 2013 for both President Kenyatta and Deputy President William Ruto related to the 2007 election aftermath, US President Barack Obama chose not to visit the country during his mid-2013 African trip. Later in the summer, Kenyatta visited China at the invitation of President Xi Jinping after a stop in Russia and not having visited the United States as president. In July 2015 Obama visited Kenya, as the first American president to visit the country while in office.\\
            \textbf{Answer:} International Criminal Court trial dates\\
            \textbf{Question:} Why did President Kenyatta and Deputy President William Ruto not visit the United States in 2013?\\
            \textbf{Scores:} fluency - 3; clarity - 3; conciseness - 3; relevance - 2.6667; consistency - 1.6667; answerability - 1.3333; answer consistency - 1
        }
        \tabularnewline
        \midrule
        \parbox{15.5cm}{
            \textbf{Example 5}\\
            \textbf{Passage:} Even before Washington returned, Dinwiddie had sent a company of 40 men under William Trent to that point, where in the early months of 1754 they began construction of a small stockaded fort. Governor Duquesne sent additional French forces under Claude-Pierre Pecaudy de Contrecœur to relieve Saint-Pierre during the same period, and Contrecœur led 500 men south from Fort Venango on April 5, 1754. When these forces arrived at the fort on April 16, Contrecœur generously allowed Trent's small company to withdraw. He purchased their construction tools to continue building what became Fort Duquesne.\\
            \textbf{Answer:} 40\\
            \textbf{Question:} How many men did Duquesne send to relieve Saint-Pierre?\\
            \textbf{Scores:} fluency - 3; clarity - 3; conciseness - 3; relevance - 3; consistency - 3; answerability - 3; answer consistency - 1
        }
        \tabularnewline
        \midrule
        \parbox{15.5cm}{
            \textbf{Example 6}\\
            \textbf{Passage:} There are fifteen fraternities and seven sororities at the University of Chicago, as well as one co-ed community service fraternity, Alpha Phi Omega. Four of the sororities are members of the National Panhellenic Conference, and ten of the fraternities form the University of Chicago Interfraternity Council. In 2002, the Associate Director of Student Activities estimated that 8–10 percent of undergraduates were members of fraternities or sororities. The student activities office has used similar figures, stating that one in ten undergraduates participate in Greek life.\\
            \textbf{Answer:} fifteen\\
            \textbf{Question:} How many fraternities are at the University of Chicago?\\
            \textbf{Scores:} fluency - 3; clarity - 3; conciseness - 3; relevance - 3; consistency - 3; answerability - 3; answer consistency - 3
        }
        \tabularnewline
        \bottomrule
    \end{tabularx}
    \caption{Annotation examples.}
    \label{tab:annotation_example}
\end{table*}

Specifically, under the fine-tuning and LoRA settings, we trained QG models separately for each base dataset. In the fine-tuning setting, for the SQuAD dataset, we utilized public fine-tuned models from Huggingface\footnote{https://huggingface.co/lmqg}, while for the HotpotQA dataset, we conducted our own model fine-tuning as there were few fine-tuned models publicly available. We set the learning rate as 1e-4, warmup steps 500, weight decay 0.01, and the max train epochs as 10 and trained the QG models on a single RTX 3090 GPU (memory limit is 24576 MiB). 
When applying LoRA, we set the learning rate as 1e-4, weight decay as 0.01, and max train epochs as 3, and trained models on an A800 GPU (memory limit is 81920MiB). 
In few-shot learning, we randomly select 8 examples to provide for the models as recommended in \cite{min-etal-2022-rethinking} that model performance does not increase much as the number of examples increases when it reaches 8. The total cost of calling GPT-3.5 and GPT-4 APIs to generate questions (800 questions) is about \$6.
\begin{table*}[t]
\centering
\small
\begin{tabularx}{\textwidth}{ll}
\toprule
\textbf{Models} & \textbf{Instructions}\\
\midrule
\parbox{3.8cm}{BART-base-finetune\\BART-large-finetune} & 
\parbox{11cm}{\{answer\} </s> \{passage\}}\\
\midrule
\parbox{3.8cm}{T5-base-finetune\\T5-large-finetune} & 
\parbox{11cm}{answer: \{answer\}  context: \{passage\}}\\
\midrule
\parbox{3.8cm}{Flan-T5-base-finetune\\Flan-T5-large-finetune\\Flan-T5-XL-LoRA\\Flan-T5-XXL-LoRA} & 
\parbox{11cm}{Generate a question based on the given answer and context. Answer: \{answer\}  Context: \{passage\}}\\
\midrule
\parbox{3.8cm}{Flan-T5-XL-fewshot\\Flan-T5-XXL-fewshot} & 
\parbox{11cm}{Generate a question based on the given passage and answer.\\ Answer: \{example\_answer\}  Context: \{example\_passage\}  Question: \{example\_question\}\\...\\Answer: \{answer\}  Context: \{passage\}  Question: }\\
\midrule
\parbox{3.8cm}{GPT-3.5-turbo-zeroshot\\GPT-4-zeroshot} & 
\parbox{11cm}{Generate a question based on the given answer and context, the generated question must be answered by the given answer.\\ Answer: \{answer\}  Context: \{passage\}  Question: }\\
\midrule
\parbox{3.8cm}{GPT-3.5-turbo-fewshot\\GPT-4-fewshot} & 
\parbox{11cm}{Generate a question based on the given answer and context, the generated question must be answered by the given answer.\\Examples:\\Answer: \{example\_answer\}  Context: \{example\_passage\}  Question: \{example\_question\}\\...\\Answer: \{answer\}  Context: \{passage\}  Question: }\\
\bottomrule
\end{tabularx}
\caption{Task instructions for different QG models.}
\label{tab:qgmodel_ins}
\end{table*}

\section{Data Statistics}
\label{apx:data_statistics}
We show some statistics of QGEval and comparison with existing benchmarks in Table~\ref{tab:statistics}. Compared to existing benchmarks, QGEval covers a broader range of dimensions, providing a more comprehensive evaluation of generated questions. Additionally, QGEval utilizes a greater variety of models, offering a more robust and thorough assessment and comparison of current QG models. 
\begin{table*}[htbp]
\centering
\small
    \begin{tabularx}{\textwidth}{p{3cm}lllp{5cm}p{1.5cm}p{2cm}}
    \toprule
    \textbf{Name} & \textbf{\#Q} & \textbf{\#P} & 
    \textbf{\#M} & \textbf{Dimensions} & \textbf{Score Scale} & \textbf{Base dataset}\\
    \midrule
    \parbox{3cm}{Q-metric \cite{nema-khapra-2018-towards}} & 3000 & --- & 0 & \parbox{5cm}{Answerability} & 1-5 & \parbox{2cm}{SQuAD, VQA,\\ WikiMovies} \rule{0pt}{3pt} \\
    \parbox{3cm}{SimQG \cite{gollapalli-ng-2022-qsts}} & 500 & 483 & 3 & \parbox{5cm}{Fluency, Relevance, Answerability,\\ Similarity} & 0-1 & \parbox{2cm}{SQuAD} \rule{0pt}{3pt} \\
    \parbox{3cm}{Quiz Design Task \cite{laban-etal-2022-quiz}} & 3164* & 7 & 7 & \parbox{6cm}{Acceptance} & 0 or 1 & \parbox{2cm}{SQuAD}\\
    \midrule
    QGEval & 3000 & 200 & 15 & \parbox{5cm}{Fluency, Clarity, Conciseness, \\Relevance, Consistency, Answerability,\\Answer Consistency} & 1-3 & \parbox{2cm}{SQuAD, \\HotpotQA}\\
    \bottomrule
    \end{tabularx}
    \caption{Detail statistics of QGEval with other benchmarks. \#Q: The number of generated questions; \#P: The number of passages; \#M: The number of QG models. *The number of questions in the Quiz Design Task does not exclude annotations from different annotators.}
    \label{tab:statistics}
\end{table*}

\section{Automatic Metrics}
\label{apx:metrics}
Detailed descriptions of the automatic metrics we evaluate are listed as follows:
\begin{itemize}
    \item \textbf{BLEU:} \cite{papineni2002bleu}, a metric that measures the number of overlapping n-grams between the generated text and a set of gold reference texts.
    \item \textbf{ROUGE:} \cite{lin-2004-rouge}, a recall-oriented metric specifically focuses on the longest common subsequence (LCS) between the generated and reference texts.
    \item \textbf{METEOR:} \cite{banarjee2005}, a metric computes an alignment between generated texts and reference texts based on the harmonic mean of unigram precision and recall.
    \item \textbf{MoverScore:} \cite{zhao-etal-2019-moverscore}, a metric measures the Earth Mover's Distance \cite{kusner2015doc} between the distributions of words in the generated text and the reference text.
    \item \textbf{BERTScore:} \cite{Zhang*2020BERTScore:}, a metric computes the semantic similarity of the generated text and reference text by leveraging contextual embeddings from BERT \cite{devlin-etal-2019-bert}.
    \item \textbf{BLEURT:} \cite{sellam-etal-2020-bleurt}, a learned metric leveraging BERT architecture to evaluate text generation.
    \item \textbf{Q-Metric:} \cite{nema-khapra-2018-towards}, a specialized metric designed for the QG task, which considers not only n-gram similarity but also the answerability of questions.
    \item \textbf{QSTS:} \cite{gollapalli-ng-2022-qsts}, a metric that utilizes the questions' types, entities, and semantic features to evaluate the similarity between questions.
    \item \textbf{BARTScore:} \cite{NEURIPS2021_e4d2b6e6}, a method that formulates evaluating generated text as a text generation task based on the BART model.
    \item \textbf{GPTScore:} \cite{fu2023gptscore}, a framework that leverages the capabilities of generative pre-trained models for evaluation. The intuition of it is similar to BARTScore.
    \item \textbf{UniEval:} \cite{zhong-etal-2022-towards}, a comprehensive framework for evaluating the generated text from multiple explainable dimensions (e.g., fluency) based on T5.
    \item \textbf{QRelScore:} \cite{wang-etal-2022-qrelscore}, a context-aware evaluation method designed for QG, incorporating word-level hierarchical matching based on BERT and sentence-level prompt-based generation techniques based on GPT-2 \cite{radford2019language}.
    \item \textbf{RQUGE:} \cite{mohammadshahi-etal-2023-rquge}, a reference-free metric that assesses the answerability of questions based on the QA model's ability to generate an answer to this question within a given context. 
\end{itemize}

We implement the above automatic metrics based on public tools or codes. Specifically, we utilize NLTK\footnote{\url{https://www.nltk.org/}} to calculate the BLEU and METEOR metrics. As for ROUGE\footnote{\url{https://pypi.org/project/rouge-score/}}, MoverScore\footnote{\url{https://pypi.org/project/moverscore/}}, and BERTScore\footnote{\url{https://pypi.org/project/bert-score/}}, we implement them with the corresponding Python packages. For the other metrics, we use their publicly available codes.

\section{More Experimental Results}
\subsection{Error Analysis of Generated Questions}
\label{apx:error_analysis}
We sampled 100 questions generated by QG models and conducted a pilot experiment to analyze the types of errors that occur in these questions. Out of these 100 questions, almost half (42\%) contain some degree of error. We find that the generated questions may: 1) be invalid questions, which are declarative sentences or incomplete; 2) be incorrectly phrased; 3) be ambiguously expressed; 4) contain unnecessary copies from the passage that hamper their conciseness; 5) contain inconsistent information with the passage; 6) ask for information not mentioned in the passage, resulting unanswerable based on the passage; 7) do not match with the answers. We present the proportion of each error type among the questions that contain errors in Table~\ref{tab:error-ratio} and show examples in Table~\ref{tab:example-error}.
\begin{table}[htbp]
    \centering
    \small
    \begin{tabularx}{\linewidth}{p{4.5cm}l}
    \toprule
        \textbf{Error Type} & \textbf{Percentage}\\
    \midrule
        Invalid Question & 2.38\%\\
        Incorrectly Phrased & 7.14\%\\
        Ambiguous & 30.95\%\\
        Unnecessary Copy from Passage & 16.67\%\\
        Inconsistent with Passage & 4.76\%\\
        Information beyond Passage & 19.05\%\\
        Mismatch with Answer & 47.62\%\\
    \bottomrule    
    \end{tabularx}
    \caption{Proportion of error types. One question may contain multiple types of errors.}
    \label{tab:error-ratio}
\end{table}
\begin{table*}[htbp]
\centering
    \small
    \begin{tabularx}{\textwidth}{ll}
		\toprule
            \textbf{Error Type} & \textbf{Example}\\
        \midrule
            \parbox{3cm}{\textbf{Invalid Question}} & 
            \parbox{12cm}{
            \textbf{Passage:} \textit{Graptopetalum (leatherpetal) is a plant genus of the family "Crassulaceae".  They are perennial succulent plants ......}\\
            \textbf{Answer:} \textit{yes}\\
            \textbf{Question:} \textit{\underline{Answer: no}} 
        } \\
        \midrule
            \parbox{3cm}{\textbf{Incorrectly Phrased}} & 
            \parbox{12cm}{
            \textbf{Passage:} \textit{...... The publication of a Taoist text inscribed with the name of Töregene Khatun, Ögedei's wife, is one of the first printed works sponsored by the Mongols......}\\
            \textbf{Answer:} \textit{Töregene Khatun}\\
            \textbf{Question:} \textit{\underline{Who was a Taoist text} inscribed with the name of gedei's wife?} 
        } \\
        \midrule
        \parbox{3cm}{\textbf{Ambiguous}} &
        \parbox{12cm}{
            \textbf{Passage:} \textit{Although most are non-aligned, some of the best known independent schools also belong to the large, long-established religious foundations, such as the Anglican Church, Uniting Church and Presbyterian Church, ......}\\
            \textbf{Answer:} \textit{Presbyterian Church}\\
            \textbf{Question:} \textit{What is \underline{another large religious foundation} that some of the best known independent schools belong to?}
        }\\ 
        \midrule
        \parbox{3cm}{\textbf{Unnecessary Copy \\from Passage}} &
        \parbox{12cm}{
            \textbf{Passage:} \textit{American burlesque is a genre of variety show.  Derived from elements of Victorian burlesque, music hall and minstrel shows, burlesque shows in America became popular in the 1860s and evolved to feature ribald comedy (lewd jokes) and female striptease.\\Brian Newman (born June 10, 1981) is an American jazz musician, singer, and trumpet player. Newman currently holds a residency at the Rose Bar at the Gramercy Park Hotel in New York City and is married to American burlesque performer Angie Pontani.}\\
            \textbf{Answer:} \textit{American burlesque}\\
            \textbf{Question:} \textit{\underline{Brian Newman (born June 10, 1981) is an American jazz musician, singer, and} \underline{trumpet player, Newman currently holds a residency at the Rose Bar at the Gramercy Park Hotel} \underline{in New York City and is married to Angie Pontani,} a performer of which genre of variety show, that became popular in the 1860s and evolved to feature ribald comedy (lewd jokes) and female striptease?}
        }\\ 
        \midrule
        \parbox{3cm}{\textbf{Inconsistent with \\Passage}} &
        \parbox{12cm}{
            \textbf{Passage:} \textit{The United States's Sculpin nuclear test series was a group of 7 nuclear tests conducted in 1990-1991.  These tests followed the "Operation Aqueduct" series and preceded the "Operation Julin" series.\\The United States's Julin nuclear test series was a group of 7 nuclear tests conducted in 1991–1992.  These tests followed the "Operation Sculpin" series, and were the last before negotiations began for the Comprehensive Test Ban Treaty.}\\
            \textbf{Answer:} \textit{Operation Aqueduct}\\
            \textbf{Question:} \textit{The United States's Sculpin nuclear test series was a group of 7 nuclear tests conducted in 1990-1991, these tests followed which series, and \underline{were the last before negotiations began for the Comprehensive Test Ban Treaty}?}
        }\\ 
        \midrule
        \parbox{3cm}{\textbf{Information beyond \\Passage}} &
        \parbox{12cm}{
            \textbf{Passage:} \textit{According to PolitiFact the top 400 richest Americans "have more wealth than half of all Americans combined." ......}\\
            \textbf{Answer:} \textit{400}\\
            \textbf{Question:} \textit{According to PolitiFact, \underline{who are the richest Americans}?}
        }\\ 
        \midrule
        \parbox{3cm}{\textbf{Mismatch with \\Answer}} &
        \parbox{12cm}{
            \textbf{Passage:} \textit{As of August 2013, Skateboardeßr Magazine is primarily a digital skateboarding publication......its Editor/Photo Editor is Jaime Owens, while the magazine's Publisher is Jamey Stone.  On August 19, 2013, the magazine's owner GrindMedia announced that the publication would cease production on October 15, 2013......}\\
            \textbf{Answer:} \textit{October 15, 2013}\\
            \textbf{Question:} \textit{\underline{What is the name of the person} who is the editor of Skateboarder Magazine?}
        }\\ 
        \bottomrule
	\end{tabularx}
\caption{Examples of errors in generated questions. Errors within questions are highlighted with underlines.}
\label{tab:example-error}
\end{table*}

\subsection{Annotation Distributions on Different Base Datasets}
\label{apx:annotation_dist}
\begin{figure*}[htbp]
    \begin{subfigure}{\linewidth}
        \includegraphics[width=\textwidth]{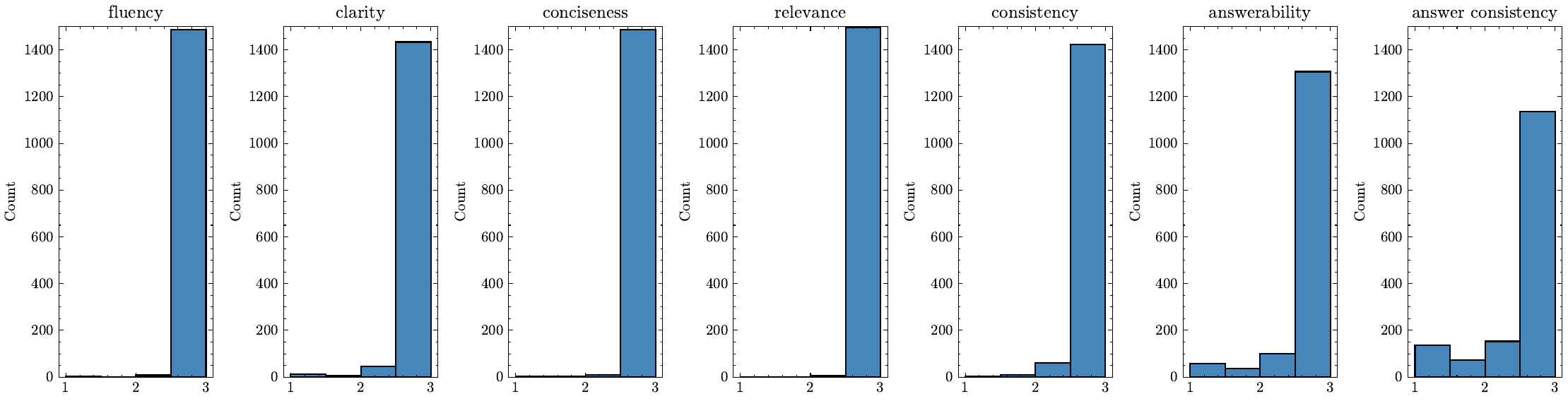}
        \subcaption{Annotation score distributions over seven dimensions on SQuAD.}
        \label{fig:annotation_dist_squad}
    \end{subfigure}
    \begin{subfigure}{\linewidth}
        \includegraphics[width=\textwidth]{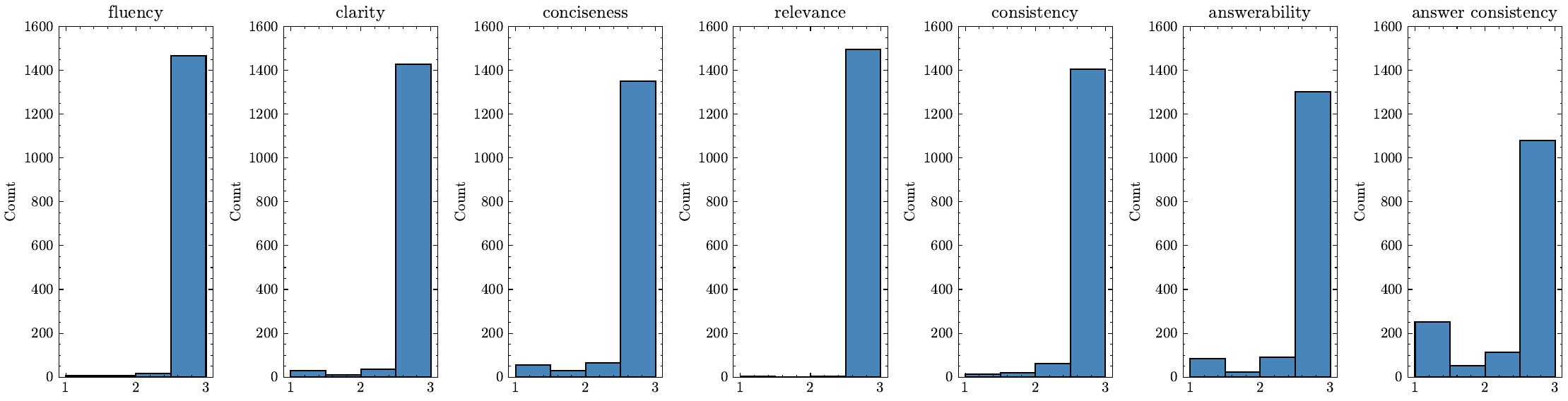}
        \caption{Annotation score distributions over seven dimensions on HotpotQA.}
        \label{fig:annotation_dist_hotpot}
    \end{subfigure}
\caption{Annotation score distributions over seven dimensions.}
\label{fig:annotation_dist}
\end{figure*}

For further insights, we also show the annotation score distribution over each dimension on the SQuAD and HotpotQA datasets in Figure~\ref{fig:annotation_dist_squad} and Figure~\ref{fig:annotation_dist_hotpot} respectively. Compared to questions generated based on SQuAD, questions generated from HotpotQA are more likely to exhibit issues in dimensions such as conciseness, answerability, and answer consistency. This tendency may arise from the fact that reference questions in HotpotQA are predominantly multi-hop questions, resulting in longer question lengths compared to those in SQuAD and posing greater difficulty in terms of answerability.

\subsection{Discriminative Power among Different Models}
\label{apx:discriminative_power}
We conducted t-tests on the annotation results across the seven dimensions, the results are shown in Table~\ref{tab:t-test_result}. The discriminative power among different models is limited; except for answer consistency, the top-five performing models fail to exhibit significant differences compared to the bottom-five models across the other six dimensions. We observe that the t-test p-value is positively correlated with the mean score difference. There is a significant difference between models with large mean score differences, whereas the differentiation between models decreases as the mean score difference decreases.

\begin{table*}[htbp]
\centering
\small
    \begin{tabularx}{\linewidth}{p{2.5cm}p{8.5cm}p{1.7cm}p{1.6cm}}
    \toprule
    \textbf{Dimensions} & \textbf{Models} & \textbf{MSDs} & \textbf{P-value}\\
    \midrule
        \multirow{3}{*}{Fluency} & GPT-3.5-turbo-zeroshot(1) vs. Flan-T5-XL-LoRA(-1) & 0.0817 & 0.0003\\
        & Flan-T5-XXL-fewshot(3) vs. Flan-T5-XXL-LoRA(-3) & 0.0483 & 0.0145\\
        & Flan-T5-large-finetune(5) vs. Flan-T5-base-finetune(-5) & 0.0183 & 0.1966\\
    \midrule
        \multirow{3}{*}{Clarity} & GPT-4-zeroshot(1) vs. Flan-T5-XL-fewshot(-1) & 0.1700 & 0.0000\\
        & GPT-3.5-turbo-zeroshot(3) vs. Flan-T5-XXL-LoRA(-3) & 0.1283 & 0.0000\\
        & Reference(5) vs. BART-base-finetune(-5) & 0.0483 & 0.0907\\
    \midrule
        \multirow{3}{*}{Conciseness} & Reference(1) vs. BART-large-finetune(-1) & 0.1700 & 0.0000\\
        & Flan-T5-XL-fewshot(3) vs. Flan-T5-XL-LoRA(-3) & 0.1050 & 0.0007\\
        & Flan-T5-base-finetune(5) vs. GPT-4-fewshot(-5) & 0.0417 & 0.1473\\
    \midrule
        \multirow{3}{*}{Relevance} & T5-base-finetune(1) vs. Flan-T5-XL-fewshot(-1) & 0.0442 & 0.0148\\
        & Flan-T5-base-finetune(3) vs. Flan-T5-XXL-LoRA(-3) & 0.0100 & 0.1073\\
        & Flan-T5-XL-LoRA(5) vs. GPT-3.5-turbo-zeroshot(-5) & 0.0050 & 0.2537\\
    \midrule
        \multirow{3}{*}{Consistency} & GPT-3.5-turbo-fewshot(1) vs. Flan-T5-XL-fewshot(-1) & 0.0467 & 0.0465\\
        & GPT-4-fewshot(3) vs. GPT-3.5-turbo-zeroshot(-3) & 0.0300 & 0.2054\\
        & BART-large-finetune(5) vs. BART-base-finetune(-5) & 0.0150 & 0.5489\\
    \midrule
        \multirow{3}{*}{Answerability} & GPT-4-fewshot(1) vs. Flan-T5-XL-fewshot(-1) & 0.2700 & 0.0000\\
        & GPT-3.5-turbo-fewshot(3) vs. BART-base-finetune(-3) & 0.1183 & 0.0139\\
        & BART-large-finetune(5) vs. Flan-T5-XL-LoRA(-5) & 0.0533 & 0.2391\\
    \midrule
        \multirow{3}{*}{\parbox{2.5cm}{Answer\\Consistency}} & GPT-4-fewshot(1) vs. GPT-3.5-turbo-zeroshot(-1) & 0.6150 & 0.0000\\
        & BART-large-finetune(3) vs. GPT-3.5-turbo-fewshot(-3) & 0.4017 & 0.0000\\
        & GPT-4-zeroshot(5) vs. BART-base-finetune(-5) & 0.1350 & 0.0301\\
    \bottomrule
    \end{tabularx}
    \caption{T-test results across seven dimensions. MSDs refer to the mean score differences.}
    \label{tab:t-test_result}
\end{table*}

\subsection{Distributions of Automatic Metrics}
\label{apx:dis_metric}
\begin{figure*}[htbp]
    \includegraphics[width=\textwidth]{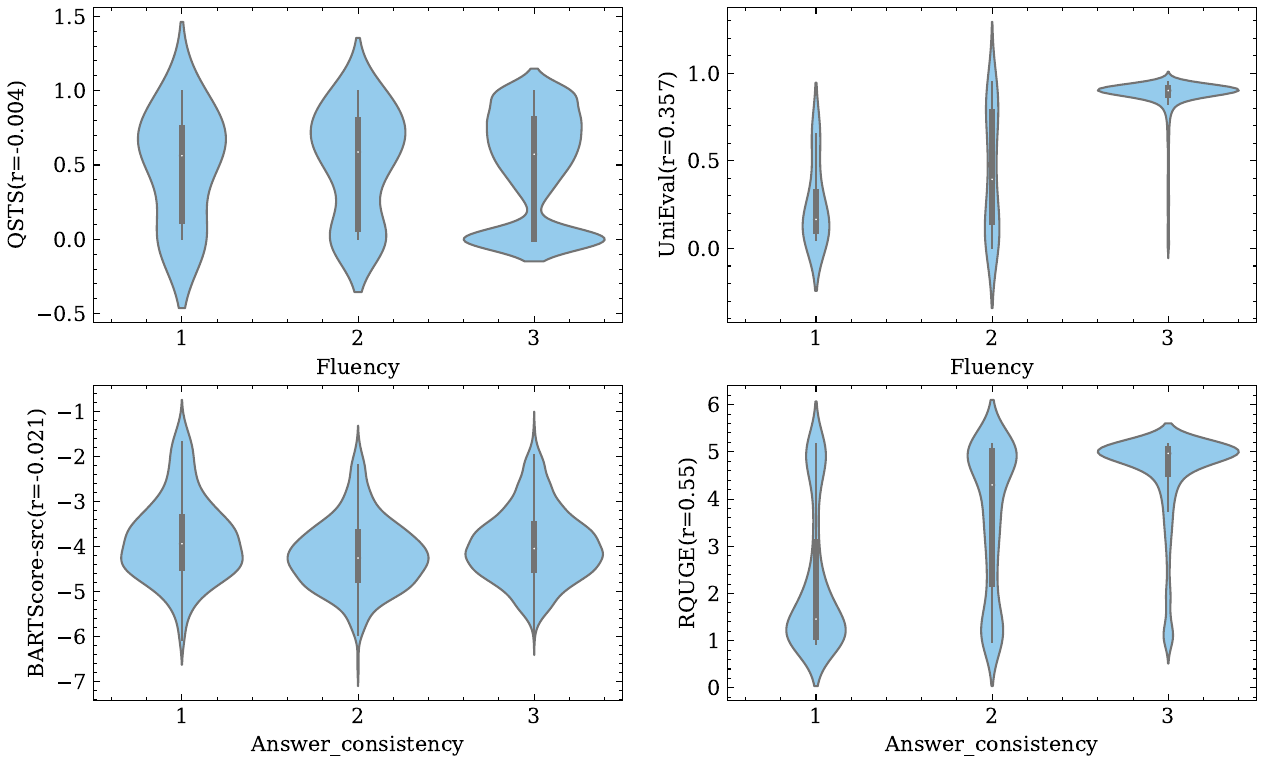}
    \caption{Distributions of automatic metrics under different human scores (1,2,3) on fluency and answer consistency.}
    \label{fig:dist_metric_more}
\end{figure*}

In Figure~\ref{fig:dist_metric_more}, we have a look at the distributions of automatic metrics under different human evaluation scores. Taking fluency (linguistic dimension) and answer consistency (task-oriented dimension) as examples, we show the distributions of the two automatic metrics that are most and least relevant to the human evaluation results. To better illustrate the distribution results in Figure~\ref{fig:dist_metric_more}, we round the human scores to the nearest integer, resulting in values of 1, 2, or 3, and then recompute the Pearson Correlations between the two automatic metrics and human scores (i.e., r in the y-axis label). 

From the figure, we observe that metrics with low correlations to human scores (e.g., QSTS on fluency and BARTScore-src on answer consistency) cannot accurately score candidates with different human scores. Metrics that achieve higher correlations with human scores (e.g., UniEval on fluency and  RQUGE on answer consistency) can correctly assign high scores to high-quality questions (human scores of 3), but they fail to distinguish accurately between questions of lower quality (human scores of 1 or 2).

\subsection{Automatic Metrics Used for Ranking QG Models}
\label{apx:metric_rank_model}
To further analyze the discriminative ability of automatic metrics across different QG models, we present the average scores of these metrics for questions generated by each model in Table~\ref{tab:metric_result}. We find that reference-based metrics appear to prefer models based on supervised training since such models excel at generating questions that are similar to the references. This type of metric faces limitations in accurately evaluating questions that are different from references, which makes them struggle to provide precise rankings of the performance of different QG models.

Reference-free metrics address the above limitations of reference-based metrics. The top three models selected based on the scores provided by these metrics partially overlap with those identified through human average scores. However, they also have constraints that result in less precise comparisons of QG models: 1) All of these metrics fail to assign high scores to the reference questions, which is a notable deficiency. 2) Metrics leveraging the generative capabilities of language models appear to exhibit a preference for questions generated by the specific model they utilize. For instance, models with the three highest GPTScore-src scores are the Flan-T5 series (Flan-T5-XL and Flan-T5-XXL), while GPTScore-src also utilizes Flan-T5-XXL as its base model. 3) Metrics designed for specific dimensions are inappropriate for overall performance comparisons across different models. For example, RQUGE is ill-suited for accurately evaluating the overall performance of QG models since it focuses only on the dimensions related to answers.

\begin{table*}[htbp]
\small
    \begin{subtable}{\textwidth}
        \centering
        \subcaption{Average scores from reference-based automatic metrics. The three highest scores for each metric are bolded. Abbreviations are as follows. B4:BLEU-4; RL:ROUGE-L; MR:METEOR; BERT:BERTScore; Mover:MoverScore; BRT:BLEURT; BART\textsubscript{ref}:BARTScore-ref; GPT\textsubscript{ref}:GPTScore-ref; QB4:Q-BLEU4.}
        \label{tab:metric_result_ref}
        \begin{tabularx}{\textwidth}{p{3.1cm}XXXXXXXXXX}
        \toprule
            \textbf{Models} & \textbf{B4} & \textbf{RL} & \textbf{MR} & \textbf{BERT} & \textbf{Mover} & \textbf{BRT} & \textbf{BART\textsubscript{ref}} & \textbf{GPT\textsubscript{ref}} & \textbf{QB4} & \textbf{QSTS}\\
        \midrule
            Reference & \textbf{1.000} & \textbf{1.000} & \textbf{0.999} & \textbf{1.000} & \textbf{1.000} & \textbf{0.979} & \textbf{-2.005} & \textbf{-0.385} & \textbf{1.000} & \textbf{1.000}\\
            BART-base-finetune & 0.162 & 0.444 & 0.428 & 0.913 & 0.638 & 0.566 & -3.502 & -1.858 & 0.374 & 0.508\\
            BART-large-finetune & 0.147 & 0.427 & 0.420 & 0.908 & 0.630 & 0.554 & -3.640 & -1.978 & 0.361 & 0.502\\
            T5-base-finetune & 0.168 & 0.467 & 0.428 & 0.914 & 0.642 & 0.559 & -3.480 & -1.891 & 0.374 & 0.504\\
            T5-large-finetune & \textbf{0.177} & \textbf{0.491} & \textbf{0.446} & \textbf{0.918} & \textbf{0.652} & \textbf{0.583} & \textbf{-3.404} & \textbf{-1.800} & \textbf{0.396} & \textbf{0.530}\\
            Flan-T5-base-finetune & \textbf{0.171} & 0.474 & 0.438 & 0.914 & 0.640 & 0.566 & -3.517 & -1.889 & 0.377 & 0.513\\
            Flan-T5-large-finetune & 0.169 & \textbf{0.482} & \textbf{0.447} & \textbf{0.917} & 0.639 & 0.572 & -3.432 & -1.824 & \textbf{0.390} & \textbf{0.528}\\
            Flan-T5-XL-LoRA & 0.160 & 0.458 & 0.429 & 0.911 & 0.637 & 0.557 & -3.560 & -1.907 & 0.370 & 0.507\\
            Flan-T5-XXL-LoRA & 0.169 & 0.474 & 0.427 & \textbf{0.917} & \textbf{0.647} & \textbf{0.577} & \textbf{-3.409} & \textbf{-1.787} & 0.384 & 0.509\\
            Flan-T5-XL-fewshot & 0.098 & 0.380 & 0.300 & 0.900 & 0.609 & 0.477 & -3.731 & -2.014 & 0.280 & 0.326\\
            Flan-T5-XXL-fewshot & 0.110 & 0.397 & 0.315 & 0.906 & 0.615 & 0.497 & -3.662 & -1.943 & 0.315 & 0.396\\
            GPT-3.5-turbo-fewshot & 0.084 & 0.330 & 0.307 & 0.891 & 0.592 & 0.474 & -4.033 & -2.134 & 0.247 & 0.383\\
            GPT-4-fewshot & 0.078 & 0.333 & 0.340 & 0.890 & 0.585 & 0.491 & -4.083 & -2.228 & 0.243 & 0.457\\
            GPT-3.5-turbo-zeroshot & 0.076 & 0.315 & 0.295 & 0.890 & 0.587 & 0.471 & -4.039 & -2.208 & 0.228 & 0.353\\
            GPT-4-zeroshot & 0.067 & 0.305 & 0.323 & 0.890 & 0.578 & 0.490 & -4.119 &-2.195 & 0.226 & 0.430\\
        \bottomrule
        \end{tabularx}
    \end{subtable}
    \begin{subtable}{\textwidth}
        \centering
        \subcaption{Average scores from reference-free automatic metrics. The three highest scores for each metric are bolded. Abbreviations are as follows. BART\textsubscript{src}:BARTScore-src; GPT\textsubscript{src}:GPTScore-src.}
        \label{tab:metric_result_src}
        \begin{tabularx}{\textwidth}{p{3.8cm}XXXXX}
        \toprule
            \textbf{Models} & \textbf{BART\textsubscript{src}} & \textbf{GPT\textsubscript{src}} & \textbf{UniEval} & \textbf{QRelScore} & \textbf{RQUGE}\\
        \midrule
            Reference & -4.558 & -1.184 & 0.881 & 0.045 & 4.140\\
            BART-base-finetune & -4.011 & -0.621 & 0.901 & 0.130 & 4.261\\
            BART-large-finetune & -3.868 & -0.632 & 0.893 & \textbf{0.155} & \textbf{4.363}\\
            T5-base-finetune & -4.041 & -0.616 & 0.892 & 0.118 & 4.273\\
            T5-large-finetune & -4.045 & -0.601 & 0.909 & 0.118 & \textbf{4.354}\\
            Flan-T5-base-finetune & -4.038 & -0.595 & 0.899 & 0.120 & 4.261\\
            Flan-T5-large-finetune & -4.025 & -0.603 & 0.908 & 0.118 & \textbf{4.355}\\
            Flan-T5-XL-LoRA & -3.962 & \textbf{-0.579} & 0.900 & 0.137 & 4.268\\
            Flan-T5-XXL-LoRA & -4.137 & \textbf{-0.566} & 0.901 & 0.104 & 4.289\\
            Flan-T5-XL-fewshot & -4.271 & -0.710 & 0.902 & 0.071 & 3.574\\
            Flan-T5-XXL-fewshot & -4.260 & \textbf{-0.441} & 0.904 & 0.078 & 3.955\\
            GPT-3.5-turbo-fewshot & \textbf{-3.652}& -0.684 & 0.908 & \textbf{0.166} & 3.555\\
            GPT-4-fewshot & \textbf{-3.548} & -0.907 & \textbf{0.938} & \textbf{0.144} & 4.240\\
            GPT-3.5-turbo-zeroshot & \textbf{-3.652} & -0.869 & \textbf{0.922} & 0.137 & 3.389\\
            GPT-4-zeroshot & \textbf{-3.638} & -1.078 & \textbf{0.931} & 0.121 & 4.216\\
        \bottomrule 
        \end{tabularx}
    \end{subtable}
\caption{Average scores of automatic metrics for questions generated by each model.}
\label{tab:metric_result}
\end{table*}

\end{document}